\pdfoutput=1
\documentclass[11pt]{article}

\usepackage[final]{acl}

\usepackage{times}
\usepackage{latexsym}

\usepackage[ruled,vlined]{algorithm2e}
\usepackage{multicol}

\usepackage[T1]{fontenc}
\usepackage[utf8]{inputenc}

\usepackage{microtype}

\usepackage{inconsolata}

\usepackage{graphicx}

\usepackage{amsmath}
\usepackage{hyperref}
\usepackage{url}
\usepackage{booktabs}
\usepackage{multirow}
\definecolor{darkblue}{rgb}{0, 0, 0.5}
\hypersetup{colorlinks=true, citecolor=darkblue, linkcolor=darkblue, urlcolor=darkblue}
\usepackage[titletoc,page]{appendix}
\usepackage{soul}
\usepackage{latexsym}

\usepackage{amsmath,amsfonts,bm}

\def\eqref#1{equation~\ref{#1}}
\def\1{\bm{1}}

\DeclareMathAlphabet{\mathsfit}{\encodingdefault}{\sfdefault}{m}{sl}
\SetMathAlphabet{\mathsfit}{bold}{\encodingdefault}{\sfdefault}{bx}{n}

\usepackage[utf8]{inputenc}
\usepackage{inconsolata}

\usepackage{listings}
\usepackage[framemethod=PStricks]{mdframed}
\usepackage{tcolorbox}
\usepackage{textcomp}
\usepackage{colortbl}
\usepackage{xcolor}
\usepackage{pifont}
\usepackage{graphicx,subfigure}
\usepackage{stfloats}
\usepackage[T1]{fontenc}

\usepackage{pythonhighlight}

\definecolor{darkgreen}{RGB}{50,100,0}
\definecolor{darkred}{RGB}{200, 0, 0}
\definecolor{lightred}{RGB}{250, 200, 200}
\definecolor{lightblue}{RGB}{210, 220, 250}
\definecolor{red}{HTML}{FFEBEE}
\definecolor{green}{HTML}{E8F5E9}
\definecolor{blue}{HTML}{E1F5FE}

\definecolor{keywords}{RGB}{255,0,90}
\definecolor{comments}{RGB}{0,0,113}

\title{Genetic Instruct: Scaling up Synthetic Generation of
Coding Instructions for Large Language Models}

\author{\textbf{Somshubra Majumdar}*, \textbf{Vahid Noroozi} \thanks{Equal contribution}, \textbf{Mehrzad Samadi, Sean Narenthiran}, \\ \textbf{Aleksander Ficek,  Wasi Uddin Ahmad, Jocelyn Huang, Jagadeesh Balam, Boris Ginsburg} \\
NVIDIA\\
\texttt{\{smajumdar,vnoroozi,msamadi,snarenthiran,aficek,}\\
\texttt{wasiuddina,jocelynh,jbalam,bginsburg\}@nvidia.com}
}

\begin{document}
\maketitle
\begin{abstract}
Large Language Models (LLMs) require high quality instruction data for effective alignment, particularly in code generation tasks where expert curated datasets are expensive to produce. We present Genetic-Instruct, a scalable algorithm for synthesizing large-scale, high quality coding instructions using evolutionary principles. Starting from a small set of seed instructions, Genetic-Instruct generates diverse and challenging instruction-code pairs by leveraging an Instructor-LLM for generation, a Coder-LLM for code synthesis, and a Judge-LLM for automatic quality evaluation. Our proposed approach is highly parallelizable and effective even with a small seed data and weaker generator models. We generated more than 7.5 million coding instructions with the proposed approach. Then we evaluated it by fine-tuning LLMs with the synthetic samples and demonstrated a significant improvement in their code generation capability compared to the other synthetic generation approaches and publicly available datasets. Our results highlight the efficiency, scalability, and generalizability of the Genetic-Instruct framework.

\end{abstract}

\section{Introduction}

Large Language Models (LLMs) have made significant progress in programming tasks and are increasingly being used as code assistants \citep{liang2024large}. To fully exploit their potential, they require alignment \citep{ouyang2022training}, which depends on paired instruction-solution examples to shape the behavior of the model. However, creating diverse and complex instructions, especially in coding domains, can be expensive due to the need for expert input. A promising alternative is to generate synthetic instructions using another LLM. Previous research shows that synthetic instructions are effective for both coding \citep{luo2024wizardcoder, wu2024inversecoder, wei2024magicoder, yu2024wavecoder} and general tasks \citep{wang-self-instruct-2023, honovich-etal-2023-unnatural, xu2024wizardlm}.

\begin{figure}[ht]
    \centering
    \vspace{-4mm}
    \includegraphics[scale=0.75]{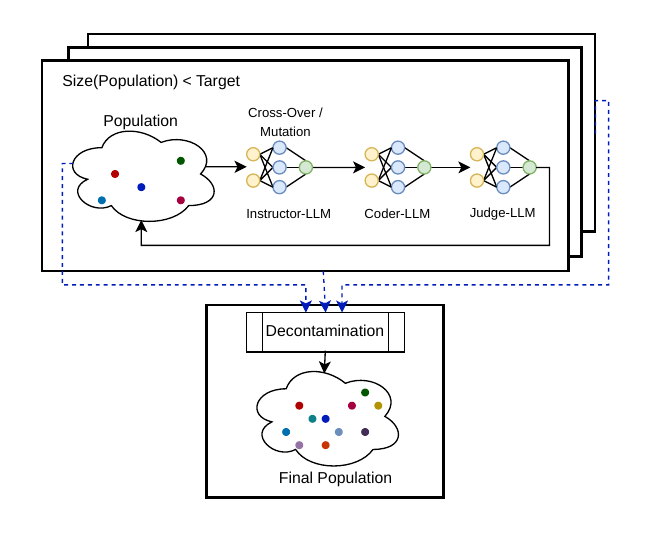}
    \vspace{-4mm}
    \caption{  
    The overall process of Genetic-Instruct across multiple parallel colonies per generation. Each colony begins with a small seed population, from which an Instructor-LLM applies crossover and mutation to create new instructions. A Coder-LLM then generates corresponding code solutions, which are evaluated by a Judge-LLM for correctness and quality. Once the target population size is reached, samples are decontaminated to form the final population.
    }
    \label{fig:Genetic-Instruct}
\end{figure}

In this paper, we introduce Genetic-Instruct, a scalable algorithm to generate synthetic coding instructions, illustrated in Figure \ref{fig:Genetic-Instruct}. Inspired by evolutionary algorithms, Genetic-Instruct starts with a small set of seed instructions and uses LLMs to generate new instruction-code pairs through two operations of crossover and mutation.

The crossover operation follows a self-instruct approach \cite{wang-self-instruct-2023}, where an LLM creates new instructions from few-shot examples, expanding the topic coverage beyond the original seeds. The crossover operator is mainly employed to enhance diversity by expanding the overall coverage of the instructions to wider domains and topics beyond the original seed instructions.

In the mutation operation, an LLM evolves a given instruction into another instruction based on some predefined rules \citep{luo2024wizardcoder}. This operation can help the generation process to increase the diversity of the instructions locally. An instruction generated by one operation is added to the pool of the seeds, and it may be used by the the operation or other in the next step. This collaborative and coupled interaction between the crossover and mutation is the main key foundation of our proposed approach. It boosts instruction diversity, which is an essential factor in the success of synthetic instruction generation.

Subsequently, another LLM generates answers, including code solutions, for the instructions. We introduce a fitness function that uses an LLM to evaluate the correctness and quality of each instruction-solution pair. Samples that pass these checks are added to the population pool, and the evolutionary process continues until the target population size is reached. Starting from a small set of seed instructions, the pool grows with newly generated synthetic instructions.

Additionally, the entire pipeline is designed for efficient parallel execution with multiple colonies of populations by running multiple instances of this process in parallel. Furthermore, this process can be repeated multiple times to generate more generations using the instructions generated from the previous round as the seed for the next generation.

Using our Genetic-Instruct algorithm, we generated a large dataset of synthetic coding instructions (more than 7.5M samples), starting from 512 seed questions. We trained LLMs on these data via supervised fine-tuning (SFT) and evaluated them on code generation benchmarks. Our work supports open-source development, avoiding any closed-source data or models. 

Models trained on our synthetic dataset achieved strong results across coding benchmarks, outperforming other instruction generation methods and also some of the existing public SFT datasets. Our experiments also show that Genetic-Instruct can produce high-quality data without requiring very strong LLMs or large seed sets. We released the dataset publicly to support open-source LLM development \footnote{\url{https://huggingface.co/datasets/nvidia/OpenCodeGeneticInstruct}}.

\section{Previous Works}

Synthetic data generation has become a practical alternative to the costly and time-consuming collection of human-curated data for LLM training. A notable method is Self-Instruct \citep{wang-self-instruct-2023}, which uses a pre-trained LLM to generate instruction-output pairs from a small seed set, then fine-tunes the base model. However, Self-Instruct focuses on general tasks, not coding. Moreover, while it can enhance the coverage of topics, the synthesized samples are often simple and not challenging enough to require additional steps to arrive at the solution.

To overcome this, Evol-Instruct \cite{xu2024wizardlm} introduces instruction mutation to create more complex and diverse tasks through meta-instructions that increase reasoning depth, impose constraints, or promote conceptual evolution. This idea was adapted to coding by WizardCoder \cite{luo2024wizardcoder}, leading to improved coding performance in models trained on such evolved instructions.

While Self-Instruct and Evol-Instruct generate instructions without using any code as seeds, another line of work \cite{yu2024wavecoder,wu2024inversecoder,wei2024magicoder} generates instructions from existing code snippets. These approaches leverage large code corpora to synthesize diverse prompts. For example, INVERSE-CODER \cite{wu2024inversecoder} generates instructions directly matched to given code, whereas OSS-Instruct \cite{wei2024magicoder} and WaveCoder \cite{yu2024wavecoder} use LLMs to create new, code-inspired instructions. However, these methods rely on large high quality and processed code samples, which may pose challenges for less common programming languages.

\newcounter{algo_counter}

\newcommand{\algolabel}[1]{\refstepcounter{algo_counter}\label{#1}}

\begin{algorithm*}[ht!]
\small
\SetAlgoLined
\SetKwInOut{Input}{Input}
\SetKwInOut{Output}{Output}

\Input{
    $N$: Number of colonies \\
    $P_{max}$: Maximum population size per colony \\
    $G_N$: Total number of generations \\
    $B_{m}$ and $B_{c}$: Number of individuals needed for mutation and cross-over respectively \\
    $P_{seed}$: Initial set of seed instructions \\
    $M_p$: Probability of selecting mutation as operator \\
    $P_{op}$: Probability distribution over the operations \{Mutation: $M_p$, Cross-over: $1-M_p$\}
}
\Output{$FinalInstructions$: Generated Synthetic Instructions for Coding Problems}

\For{$g \gets 1$ \textbf{to} $G_N$}{
    Run $N$ colonies in parallel\;
    \ForEach{colony}{
        Initialize $P_{pool} \gets P_{seed}$\;
        \While{len($P_{pool}$) $<$ $P_{max}$}{
            $OP \gets$ Choose an operation from $P_{op}$\;
            $Candidates \gets$ Select a subset of $B_{m}$ or $B_{c}$ individuals from $P_{seed}$ randomly based on the selected operation\;
            $NewQuestions \gets InstructorLLM(Candidates, OP)$\;
            $FilteredQuestions \gets FilterQuestions(NewQuestions)$\;
            $GeneratedInstructions \gets CoderLLM(FilteredQuestions)$\;
            $ValidatedInstructions \gets ValidateCode(GeneratedInstructions)$\;
            $NewInstructions \gets JudgeLLM(ValidatedInstructions)$\;
            $P_{pool} \gets P_{pool} \cup NewInstructions$\;
        }
    }
    $G_g \gets$ Aggregate all $P_{pool}$ from $N$ colonies\;
}

$AggInstructions \gets$ Aggregate all $G_g$, for $g \in [1, G_n]$\;
$FinalInstructions \gets Decontaminate(AggInstructions)$\;

  \caption{Pseudo-code for the Genetic-Instruct Algorithm}
  \label{algo:genetic_instruct_algo}
\end{algorithm*}

\section{Genetic-Instruct}

We introduce Genetic-Instruct, an algorithm inspired by the population-based genetic algorithms \cite{golberg1989genetic}. This algorithm employs the two primary evolutionary operations of mutation and crossover to evolve and generate new generations from an initial population. The initial population, termed Generation 0, comprises a limited set of high-quality seed instructions. These seed instructions undergo a series of evolutionary operations, mainly mutation, crossover and selection, to transform them into new instructions. All the operations are executed by leveraging LLMs and enhancing their output with in-context learning.

The whole process of Genetic-instruct is as follows. At each step, from the instruction set of the initial population (seed population), we randomly select a batch of instructions with replacement. The LLM responsible for instruction generation (called Instructor-LLM) is employed to synthetize the new instructions based on a selected operation. Upon generating a new instruction, another LLM, referred to as the Coder-LLM, is tasked with producing the code corresponding to this new instruction. The newly generated instruction and its associated code constitute a new coding instruction, which can be utilized for training. However, there may be instances where the generated code does not fully address the provided question, or the question itself may be poorly formulated. To assess the quality of the new coding instruction, we employ another LLM, termed the Judge-LLM, to evaluate the correctness of the instruction and its code. If a sample passes this quality assessment, it is added to the pool of instructions and may be selected as the seed instruction for the next batch of synthesized samples. The entire process is iterated multiple times to synthesize samples until the desired population size is achieved. This resulting population is then labeled as a generation, and the entire pipeline can be repeated by considering this generation as the initial population for the next generation. 

Subsequently, a decontamination process is applied to minimize risk of contaminated instructions in the training data. The complete pipeline is illustrated in Figure \ref{fig:Genetic-Instruct} for one generation, and the procedure for the whole algorithm is detailed in Algorithm \ref{algo:genetic_instruct_algo}. In the following, each step is explained in detail.

\subsection{Mutation Operation}
The mutation operation is inspired by an adaptation of the Evol-Instruct algorithm, as devised by \cite{xu2024wizardlm}, and further extended by WizardCoder \citep{luo2024wizardcoder} to facilitate instruction generation for code models. Evol-Instruct evolves an instruction into another using an LLM based on predefined tasks. For a sample selected for mutation, we randomly choose one of the five tasks defined and apply the mutation to generate a new instruction. We employ the same five tasks introduced by \cite{luo2024wizardcoder}, with minor prompt modifications to suit our Instructor-LLM. Details on the mutation prompts are provided in Appendix \ref{appendix:B}.

\subsection{Crossover Operation}
The crossover operation in Genetic-Instruct is influenced by the concepts introduced in Self-Instruct \citep{wang-self-instruct-2023} and Unnatural Instructions \citep{honovich-etal-2023-unnatural}. It inspires from multiple instructions and employs the Instructor-LLM to generate new populations from the provided few-shot example instructions. To enhance the efficiency of the crossover operation, we provide multiple seed instructions and request the model to generate multiple diverse new instructions based on the provided examples in a single Instructor-LLM call. The prompt for the crossover operation is depicted in Appendix \ref{appendix:C}.

\subsection{Code Generation}
After the Instructor-LLM generates a batch of new instructions, they are passed to the Coder-LLM to generate the corresponding code solutions. The Coder-LLM should be proficient in coding tasks to ensure the generation of high-quality solutions. However, some generated code may not be parseable or compilable. Therefore, we filter out solutions whose code segments cannot be parsed by the corresponding language's parser/compiler. While determining the correctness of code by execution is the ideal case, it is challenging due to various factors, such as language constraints, missing dependencies, or having to integrate the current solution into a much larger codebase that may not be available in its entirety. The prompt used in this step is illustrated in Appendix \ref{appendix:D}.

\subsection{Fitness Function}
Simple post-processing, such as rejecting all samples that don't pass the Abstract Syntax Tree checks, is applied to filter out incorrect instructions. Then, they are scored using a fitness function in order to discard candidates that have low quality. We employ a Judge-LLM to assign a binary score indicating whether a candidate code solution meets the minimum requirements. The Judge-LLM is provided with an instruction and its code solution to determine the correctness of the instruction and its corresponding solution. To enhance the performance, we employ techniques such as in-context learning with few-shot examples and Chain-of-Thought \citep{weiChainOfThought2022} prompting to making a better decision. The prompt for the Judge-LLM is depicted in Appendix \ref{appendix:E}.

\subsection{Scaling Up the Process}
An advantage of genetic algorithms is their inherent capacity for parallelization. When utilizing computationally intensive LLMs for sample generation, it is crucial to leverage this parallel structure. We execute multiple colonies of populations in parallel processes and synchronize them periodically. These colonies are evolved and populated independently, starting from the same seed population. Upon reaching the desired size, the colonies are merged into a single population and called a generation. Additionally, to improve the diversity, we make sure that seed examples selected to be used in a batch are all different.

\subsection{LLM Decontamination}
To prevent any evaluation benchmark questions from leaking into our training samples, we adopted the decontamination methodology proposed by \citet{yang2023rethinking}, which involves two primary stages. First, for each synthesized question, we performed an embedding-based similarity search using a Sentence Transformer \cite{reimers2020multilingual} model to identify the most similar test example from all benchmark datasets. Second, we constructed question pairs by matching each synthesized question with its most similar test example. An LLM, specifically \texttt{Meta-Llama-3-70B-Instruct}, was then employed to evaluate whether any of these pairs constituted a paraphrase (details on the prompt are provided in Appendix ~\ref{appendix:F}). 

To control for potential positional bias in the LLM's paraphrase detection, we generated two pairs for each match: one where the synthesized question appeared first and another where the test set question was presented first \citep{toshniwal2024openmathinstruct2acceleratingaimath}. If any of these pairs were determined to be similar by the LLM, the synthesized question was removed.

\section{Experiments}
We fine-tune the base LLM models using supervised fine-tuning (SFT) to evaluate the effectiveness of a given instruction set. In all experiments, the models are evaluated on four benchmark datasets: HumanEval (HE) \citep{chen2021evaluating}, MBPP \citep{odena2021programsynth}, HumanEval+ (HE+), and MBPP+ \citep{evalplus}. The MBPP+ and HumanEval+ datasets, part of the EvalPlus benchmark, are extensions of the original MBPP and HumanEval test sets, respectively. These extensions include additional test cases designed to ensure the correctness and accuracy of the generated code. The prompts used for the evaluation benchmarks are provided in Appendix \ref{appendix:G}. All code evaluations are conducted using greedy decoding. Prior to SFT training, all training datasets undergo a decontamination process.

We use 512 samples from the Tiger-Leetcode collection~\citep{tigerbotkaggle2023} as the initial population in most experiments. This collection serves as the seed dataset for the first generation and consists of interview-style coding questions. Throughout all experiments, we employ the same generation models as Instructor-LLM, Coder-LLM, and Judge-LLM. Since our evaluation focuses exclusively on Python coding benchmarks, we constrain the generated solutions to Python by instructing the models to produce only questions that can be answered with Python code. After code is generated by Coder-LLM, we verify its syntactic correctness using Python's \texttt{ast} package, regardless of its executability, to ensure the structural validity of the generated code.

\subsection{Experimental Settings}
We used the AdamW optimizer \citep{Kingma15adam} for all supervised fine-tuning (SFT) experiments, with a learning rate of 5e-6 decaying to 5e-7 over three epochs, following a cosine annealing schedule \cite{loshchilov2022sgdr}. All models were trained using tensor parallelism and BF16 precision to accelerate the training process. Experiments were conducted using the NeMo framework \cite{NeMo} and NeMo Aligner \cite{NeMo-Aligner}.

For high-throughput inference with large effective batch sizes, we used vLLM \citep{kwon2023vllm} as the inference engine. Nucleus sampling \citep{Holtzman2020nucleus_sampling} was employed for decoding, with a temperature of 1.2 for Instructor-LLM, and 1.0 for both Coder-LLM and Judge-LLM. To improve GPU utilization and speed up generation, we ran 20 colonies in parallel for each generation step. A maximum sequence length of 1024 tokens was set across all LLMs to optimize generation speed and memory usage.

For Genetic-Instruct, the mutation probability ($M_p$) was set to 0.5 by default. During the mutation operation, a batch size of 100 ($B_m$) was used, while the crossover operation used a batch size of 10 ($B_c$). These values were chosen based on our observation that, the model generates approximately 10 unique instructions per generation on average, aiming to maintain a consistent number of generated samples per batch. In the crossover operation, Instructor-LLM used 3-shot in-context learning and was prompted to generate up to 20 new instructions.

\begin{figure}[t]
    \centering
    \includegraphics[scale=0.60]{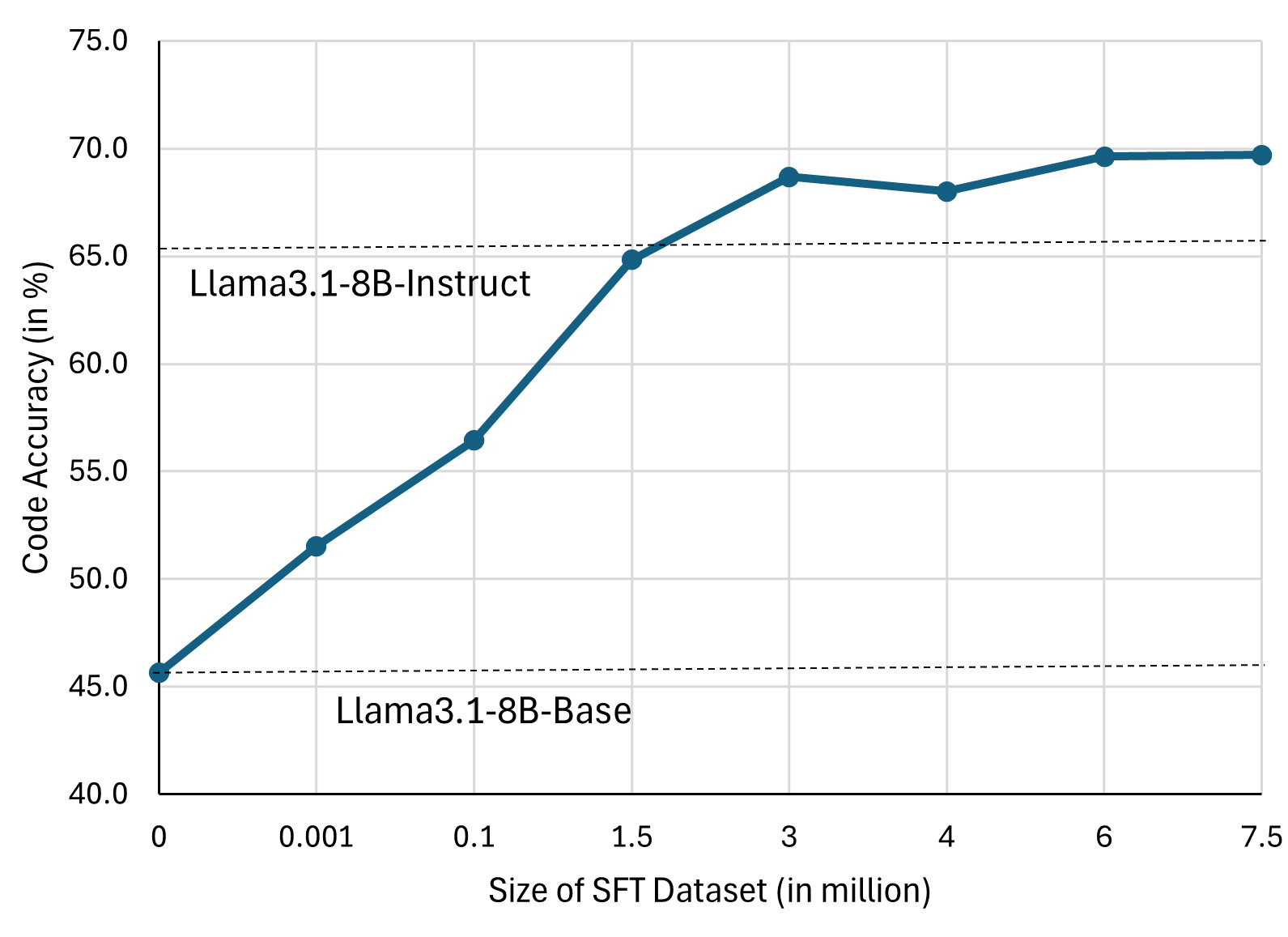}
    \caption{The accuracy of Llama-3.1-8B trained on different data sizes. Code accuracy is calculated as the average of the model's accuracy on all the four benchmarks. With scaling up the synthetic, accuracy improves but starts to show diminishing improvements later.}
    \label{fig:experiments}
\end{figure}

\begin{table*}[ht]
    \setlength{\tabcolsep}{3pt}
    \centering
    \small
    \resizebox{1.0\linewidth}{!} {%
    \def\arraystretch{1.2}%
    \begin{tabular}{lc|cc|cc|c}
        \hline
        Generation Algorithm/Dataset & Data Size & MBPP & MBPP+ & HumanEval & HumanEval+ & Average \\
        \hline
        Llama 3.1 8B Instruct   & - & 73.0&	62.7&	66.5&	61.6&	65.9 \\
        \hline
        Genetic Instruct        & 7.5M    & \textbf{79.9} &	\textbf{69.1} &	\textbf{66.5} &	\textbf{63.4} &	\textbf{69.7} \\
        Genetic Instruct        & 4M    & 76.5	 &66.9	&65.9	&62.8 &	68.0 \\
        \hline
        \rowcolor{gray!20}\multicolumn{7}{c}{Alternative Synthetic Data Generation Methods} \\
        \hline    
        WizardCoder             & 4M    & 72.8&	62.4&	65.9&	61.6&	65.7 \\
        Self-Instruct           &4M & 74.9&	66.7&	64.6&	61.0&	66.8 \\
        OSS-Instruct            &4M & 73.3&	61.4&	62.2&	58.5&	63.9 \\
        INVERSE-INSTRUCT        &4M & 59.8&	49.2&	29.3&	26.2&	41.1 \\
        \hline
        \rowcolor{gray!20}\multicolumn{7}{c}{Public Datasets} \\
        \hline
        Code Parrot Apps        & 5k    & 39.7  & 34.7  & 29.9  & 28.1  & 33.1  \\
        TACO                    & 25K   & 47.1  & 40.2  & 31.1  & 27.4  & 36.5  \\
        OpenCoder Stage 1       & 1M    & 67.2  & 57.1  & 66.5  & 61.0  & 62.9  \\
        OpenCoder Stage 2       & 170K  & 67.5  & 61.1  & 58.5  & 56.1  & 60.8  \\
        Code Alpaca             & 20K   & 31.8  & 26.7  & 24.4  & 20.7  & 25.9  \\ \hline
    \end{tabular}
    }
\caption{Comparison of Genetic-Instruct with other data generation algorithms and datasets. Average of the accuracies on all the benchmarks are also reported.}
\label{tab:main_results}
\end{table*}

\subsection{Performance Evaluation}
In this section, we evaluate the effectiveness of our proposed approach for generating synthetic supervised fine-tuning (SFT) samples aimed at enhancing the coding capabilities of LLMs. We used Llama3.1-8B-Base \citep{Llama3Herd2024} as the base model and employed Mixtral-8x22B \citep{jiang2024mixtral} as the Instructor-LLM, Coder-LLM, and Judge-LLM.

Figure~\ref{fig:experiments} illustrates the relationship between the size of the SFT dataset generated by Genetic-Instruct and coding accuracy. Coding accuracy is computed as the average model performance across all four benchmarks. We generated synthetic instructions across six generations, each consisting of approximately 1.5 million samples, totaling around 7.8 million samples. The results show a clear upward trend, where increasing the dataset size leads to significant improvements in accuracy. Notably, models trained on more than 3 million samples outperform the Llama3.1-8B-Instruct model. Starting from a baseline accuracy of approximately 45\%, the Llama3.1-8B-Base model shows consistent improvement as the dataset grows, demonstrating the scalability and effectiveness of our synthetic data generation strategy. However, beyond approximately 6 million samples, the accuracy gains begin to plateau, indicating diminishing returns.

To show the effectiveness of Genetic-Instruct compared to other approaches, we evaluated the samples generated by Genetic-Instruct with some other baseline approaches which are designed for generating synthetic SFT data for coding problems. To make the comparisons fair, we re-implemented all the baseline approaches and performed the comparisons with the same generator model, seed population, base model for SFT, and size of training data. We did not rely on the results reported in the original papers, as each one used different generation models, seed populations, base models and benchmarks. Among these baselines, WizardCoder and Self-Instruct follow a similar paradigm to ours, using a collection of coding questions to expand into a larger instruction set. In contrast, OSS-Instruct \cite{wei2024magicoder} and INVERSE-INSTRUCT \cite{wu2024inversecoder} generate instructions from a large set of real code snippets.

For OSS-Instruct and INVERSE-INSTRUCT, we used around 1.4M Python functions extracted from Stack v2 \cite{lozhkov2024starcoder} as the seed population, following the seed collection procedure adopted in \citet{wei2024selfcodealign}, while for the rest of the baselines we used Tiger-Leetcode. The same number of samples are generated by each one of the approaches with three generations. Extra samples from the last generation are dropped randomly to make all the sizes exactly 4M. The results of 5 generations (7.5M) are also reported for Genetic-Instruct. We also evaluated some of the publicly available coding instruction datasets: Apps~\citep{hendrycksapps2021}, TACO~\citep{li2023taco}, and OpenCoder~\citep{Huang2024OpenCoderTO}. All the results are presented in Table \ref{tab:main_results}. 

\begin{table}[b]
\resizebox{1.0\linewidth}{!} {%
    \centering
    %\small
    \begin{tabular}{l|cc|cc|c}
    \hline
    \textbf{Generation Algorithm} &  \textbf{MBPP} & \textbf{MBPP+} & \textbf{HE} & \textbf{HE+} & \textbf{Avg} \\
    \hline
    Cross-Over Only     & 74.9	& 66.7&	64.6&	61.0&	66.8 \\
    Mutation Only     & 73.3 &	64.0& \textbf{66.5}&	\textbf{62.8}&	66.6\\
    Genetic Instruct     & \textbf{76.5}	 &\textbf{66.9}	&65.9	&\textbf{62.8} &	\textbf{68.0} \\    
    \hline
\end{tabular}
}
\caption{ Comparing the effectiveness of different operations in the Genetic-Instruct algorithm. We generate 4 million samples for each experiment and used Llama 3.1 8B Base as the base model. 
}
\label{tab:algorithm_results}
\end{table}

For OSS-Instruct and INVERSE-INSTRUCT, we used around 1.4M Python functions extracted from Stack v2 \cite{lozhkov2024starcoder} as the seed population, following the procedure outlined in \citet{wei2024selfcodealign}. For the remaining baselines, we used Tiger-Leetcode as the seed dataset. For each approach, we generated the same number of samples over three generations, and any extra samples from the final generation were randomly discarded to standardize the dataset size to 4 million. For Genetic-Instruct, we also report results with five generations (more than 7.5M samples). Additionally, we evaluated models fine-tuned on publicly available coding instruction datasets: Apps~\citep{hendrycksapps2021}, TACO~\citep{li2023taco}, and OpenCoder~\citep{Huang2024OpenCoderTO}. The results are summarized in Table~\ref{tab:main_results}.

The results clearly highlight the superior performance of Genetic-Instruct across multiple evaluation metrics. Models trained on data generated by our method consistently outperform those trained on all baseline approaches and public datasets. In particular, our five-generation dataset achieves a significantly higher average accuracy of 69.7\% compared to the best-performing public dataset, OpenCoder Stage 1, at 62.9\%. Even our smaller dataset (4M) achieves an average of 68.0\%, further underscoring the effectiveness and efficiency of our approach.

\begin{table*}[t]
\small
\centering
\resizebox{1.0\linewidth}{!} {%
    \begin{tabular}{l|l|cc|cc|c} 
    \toprule
     \textbf{Base Model} & \textbf{Generation Model} & \textbf{MBPP} & \textbf{MBPP+} & 
    \textbf{HumanEval} & \textbf{HumanEval+} & \textbf{Average} \\
    \midrule
    \multirow{4}{*}{Llama3.1 8B}  & Mixtral 8x22B  & 72.8	&64.0&	62.8&	59.8&	64.8 \\
     & Mixtral 8x7B  & 66.7&	57.7&	52.4&	49.4&	56.5 \\
     & Qwen 32B   & \textbf{74.6}&	\textbf{65.1}&	65.2&	62.8&	\textbf{66.9} \\
     & Qwen 7B  & 72.2&	61.9&	\textbf{67.7}&	\textbf{64.0}&	66.5\\
   \midrule
    \midrule
    \multirow{4}{*}{Qwen2.5 7B}  & Mixtral 8x22B  & 79.1&	67.2&	72.6&	65.9&	71.2\\
     & Mixtral 8x7B  & 78.8&	67.2&	72.0&	65.2&	70.8 \\
     & Qwen 32B   &  \textbf{82.0}&	\textbf{72.8}&	79.3&	\textbf{75.0}&	\textbf{77.3} \\
     & Qwen 7B  & 81.2&	69.6&	\textbf{81.1}&	\textbf{75.0}	&76.7\\
    \bottomrule
    \end{tabular}
}
\caption{Ablation study on the effect of the generator model on the quality of the data generation. Average of the accuracies on all the benchmarks are also reported.}
\label{tab:generator_ablation}
\end{table*}

\subsection{Ablation Study}
\label{sec:ablation_algo}

In this ablation study, we assess the impact of mutation and crossover operations in Genetic-Instruct on the quality of generated data. We compare three setups: \textit{Crossover-Only}, where only the crossover operation is used during data generation; \textit{Mutation-Only}, where only the mutation operation is applied; and the full \textit{Genetic-Instruct} approach, which employs both. 

For each setup, we generated three generations totaling 4 million samples and fine-tuned a Llama3.1-8B Base model to evaluate downstream performance. This setup allows us to assess the individual and combined impact of these genetic operators on downstream model performance. Mutation-Only resembles WizardCoder conceptually, but with a key distinction: it updates the evolving seed pool with newly generated samples, unlike WizardCoder, which evolves only the initial seeds.

As shown in Table~\ref{tab:algorithm_results}, combining both operations yields the highest average accuracy across all benchmarks, confirming their complementary benefits. While Mutation-Only slightly outperforms the full approach on the HE benchmark, these findings suggest that while both operations individually contribute to improved performance, and their synergistic combination in Genetic-Instruct yields the most substantial overall gains in coding capability.

\subsection{Influence of the Generator Model}
Table~\ref{tab:generator_ablation} presents an ablation study evaluating the impact of different generator models on the quality of the synthetic data. We generated 1.5 million samples for each experiment with different generation models and then trained Llama3.1-8B-Base and Qwen2.5-7B-Base on them. The results indicate that the Qwen models \cite{yang2024qwen2} outperform the Mixtral family across most benchmarks, highlighting that stronger LLMs tend to produce higher-quality synthetic data.

Interestingly, Qwen-7B performs closely to Qwen-32B, suggesting that even a smaller model within the Qwen family is capable of generating high-quality training data. These findings imply that while the strength of the generator plays a key role in data quality, relatively smaller LLMs can still yield competitive performance, offering a more cost-effective alternative for synthetic data generation.

\section{Conclusion}
We introduced Genetic-Instruct, a novel algorithm inspired by evolutionary principles to generate synthetic coding instructions for LLMs. Genetic-Instruct is specifically designed to support parallel generation, making it a scalable solution for synthetic data creation. We benchmarked our approach against several baseline methods and publicly available datasets, and the results consistently demonstrated its effectiveness in improving performance on code generation tasks. Also in our ablation studies, we demonstrated the effectiveness of combining the two main operations to achieve the best performance. We publicly released the 7.5M synthetic instruction-solution dataset to facilitate the development of open source LLMs.

\bibliography{bibfile}

\newpage
\clearpage
\onecolumn

\appendix

\newpage
\section{Mutation Prompts}
\begin{figure}[h]
  \centering

\begin{tcolorbox}[title={Mutation Prompt}, colback=red!0, left=2pt,right=2pt,top=2pt,bottom=2pt]

{ %\footnotesize
Please increase the difficulty of the given programming test question a bit. Do not provide any hints, solutions or outputs. Only one new instruction is allowed.

\vspace{0.1cm}

You can increase the difficulty using, but not limited to, the following methods:

\{method\}

\vspace{0.1cm}

Original Instruction:

\{instruction\}

\vspace{0.1cm}

New Instruction:
}
\end{tcolorbox}

\begin{tcolorbox}[title={Operation: Constraint}, colback=blue!100, left=1pt,right=1pt,top=2pt,bottom=2pt]   
{\footnotesize
Rewrite the original instruction, adding new constraints and requirements, with approximately 10 additional words.
}
\end{tcolorbox}

\begin{tcolorbox}[title={Operation: Deepening}, colback=blue!100, left=1pt,right=1pt,top=2pt,bottom=2pt]   
{\footnotesize
Write the original instruction. Then, replace a commonly used requirement in the programming task with a less common and more specific.
}
\end{tcolorbox}

\begin{tcolorbox}[title={Operation: Erroneous Code}, colback=blue!100, left=1pt,right=1pt,top=2pt,bottom=2pt]   
{\footnotesize
Write the original instruction. Then provide a piece of wrong python code as a reference solution to increase misdirection.

Your wrong reference solution should start with "Reference Solution (Wrong)", marked in ``` blocks. 

Finally, ask to write the correct solution for the instruction. Do NOT provide the correct solution.
}
\end{tcolorbox}

\begin{tcolorbox}[title={Operation: Reasoning}, colback=blue!100, left=1pt,right=1pt,top=2pt,bottom=2pt]   
{\footnotesize
Write the original instruction after the new instruction. Then, if the original instruction can be solved with only a few logical steps, please add more reasoning steps after the original instruction. 

Do NOT provide any reason or explanation.
}
\end{tcolorbox}

\begin{tcolorbox}[title={Operation: Task Complexity}, colback=blue!100, left=1pt,right=1pt,top=2pt,bottom=2pt]   
{\footnotesize
Write the original instruction after the new instruction. Then propose higher time or space complexity requirements, but please refrain from doing so frequently.
}
\end{tcolorbox}

  \caption{Prompt template for mutation operation}
  \label{fig:mutation_prompt}
  \vspace{-0.2in}
\end{figure}

\label{appendix:B}

\newpage
\section{Crossover Prompt}
\begin{figure}[h]
  \centering

\begin{tcolorbox}[title={Crossover Prompt}, colback=red!0, left=2pt,right=2pt,top=2pt,bottom=2pt]
{ %\footnotesize

You are asked to come up with a set of 20 diverse code generation task instructions. These task instructions will be given to a GPT model and we will evaluate the GPT model for completing the instructions.

\vspace{0.1cm}

Here are the requirements:

1. Try not to repeat the verb for each instruction to maximize diversity.

2. The language used for the instruction also should be diverse. For example, you should combine questions with imperative instructions.

3. The type of instructions should be diverse. The list should include diverse types of programming tasks like open-ended generation, classification, editing, optimization etc.

4. A GPT language model should be able to complete the instruction. 

5. The instructions should be in English.

6. The instructions should at least 1 to 2 sentences long. Either an imperative sentence or a question is permitted.

7. You should generate an appropriate input to the instruction. The input field should contain a specific example provided for the instruction. It should involve realistic data and should not contain simple placeholders. The input should provide substantial content to make the instruction challenging but should ideally not exceed 100 words.

8. Not all instructions require input. For example, when a instruction asks about some general information, "write a program to load a file.", it is not necessary to provide a specific context. In this case, we simply put "\textlangle noinput\textrangle" in the input field.

9. The output should be an appropriate response to the instruction and the input.

10. All tasks should be coding or programming-related.

\vspace{0.1cm}
List of 20 tasks:

}
\end{tcolorbox}

\begin{tcolorbox}[title={Few-Shot Examples}, colback=blue!100, left=1pt,right=1pt,top=2pt,bottom=2pt]   
{\footnotesize

\#\#\#
\vspace{0.1cm}

1. Instruction: Convert a Binary Search Tree to a sorted Circular Doubly-Linked List in place. You can think of the left and right pointers as synonymous to the predecessor and successor pointers in a doubly-linked list. For a circular doubly linked list, the predecessor of the first element is the last element, and the successor of the last element is the first element. We want to do the transformation in place. After the transformation, the left pointer of the tree node should point to its predecessor, and the right pointer should point to its successor. You should return the pointer to the smallest element of the linked list.

1. Input: root = 4,2,5,1,3

\#\#\#

2. Instruction: $\cdots$

$\cdots$

\#\#\#

3. Instruction: 

}
\end{tcolorbox}

  \caption{Prompt template for the crossover operation with few-shot in-context learning}
  \label{fig:crossover_prompt}
  \vspace{-0.2in}
\end{figure}

\label{appendix:C}

\newpage
\section{Prompts for Coder-LLM}
\begin{figure}[h]
  \centering

\begin{tcolorbox}[title={Python Code Generation Prompt}, colback=red!0, left=2pt,right=2pt,top=2pt,bottom=2pt]

{ %\footnotesize

You are an expert in Python coding. Using only Python code, write the correct solution that answers the given coding problem.

\{instruction\}

Answer:

}
\end{tcolorbox}

  \caption{Prompt template for code Generation with Coder-LLM}
  \label{fig:coder_llm_prompt}
  \vspace{-0.2in}
\end{figure}

\label{appendix:D}

\section{Fitness Prompt for Judge-LLM}
\begin{figure}[h]
  \centering

\begin{tcolorbox}[title={Fitness Prompt}, colback=red!0, left=2pt,right=2pt,top=2pt,bottom=2pt]

{

You are an expert python programmer. 

Below is a question and code solution. Decide if the solution follows the below criteria and give a final Yes/No, and place it in the \textlangle judge\textrangle\textlangle/judge\textrangle \, tags.

Only look at the function generated, not any examples/print statements etc. Just the core logic.

Please first briefly describe your reasoning (in less than 30 words), and then write Decision: \textbackslash\textbackslash boxed\{Yes or No\} in your last line.

\vspace{0.2cm}

Criteria:

1. \textlangle llm-code\textrangle\textlangle/llm-code\textrangle \, contains a code solution in any programming language.

2. If the code was executed with the proper libraries imported and correct inputs, it would execute without error.

3. Given the question, the code solution seems to answer the problem if it was to be used correctly.

4. The code solution provides an elegant solution to the problem and doesn't seem overly complicated.

}
\end{tcolorbox}

\begin{tcolorbox}[title={Few-Shot Examples}, colback=blue!100, left=2pt,right=2pt,top=2pt,bottom=2pt]

{

Question: \{instruction\}

\textlangle llm-code\textrangle

\{code\}

\textlangle/llm-code\textrangle

\textlangle judge\textrangle

\{reason\} 

Score: \textbackslash\textbackslash boxed\{score\}.

\textlangle/judge\textrangle

}
\end{tcolorbox}
  \caption{Prompt template for code quality judgement with Judge-LLM}
  \label{fig:judge_llm_prompt}
  \vspace{-0.2in}
\end{figure}

\label{appendix:E}

\newpage
\section{Decontamination Prompt}
\begin{figure}[h]
  \centering
\begin{tcolorbox}[title={Prompt Template for Contamination Detection}, colback=red!0, left=2pt,right=2pt,top=2pt,bottom=2pt]

{
 Help me determine if the following two coding problems are the same.
  \vspace{0.3cm}

  First problem: \{instruction 1\}
  \vspace{0.3cm}

  Second problem: \{instruction 2\}
  \vspace{0.3cm}

  \vspace{0.3cm}

  Disregard the names and minor changes in word order that appear within.
  If the two problems are very similar and if they produce the same answer, we consider them to be the same problem.
  Respond with only "True" (problems are the same) or "False" (problems are different). Do not respond with anything else.
}
\end{tcolorbox}
  \caption{Prompt template for checking contamination}
  \label{fig:decontamination_prompt}
  \vspace{-0.2in}
\end{figure}
\label{appendix:F}

\newpage
\section{Evaluation Prompts}
\begin{figure}[h]
  \centering

\begin{tcolorbox}[title={Evaluation Prompt Template for MBPP and MBPP+}, colback=red!0, left=2pt,right=2pt,top=2pt,bottom=2pt]

{
Here is a problem for which you need to generate code:

\vspace{0.3cm}
\{instruction\}
\vspace{0.3cm}

Please continue to complete the code with python programming language. 

\vspace{0.3cm}
The solution should be in the following format:

\vspace{0.3cm}

  ```python

\vspace{0.3cm}

  \# Your code here

\vspace{0.3cm}

  ```

\vspace{0.3cm}

  Do not generate any tests. Your function should have the same name as the function in the assert statement. 
}
\end{tcolorbox}

  \caption{Prompt template for code evaluation on MBPP and MBPP+}
  \label{fig:mbpp_prompt}
  \vspace{-0.2in}
\end{figure}

\begin{figure}[h]
  \centering

\begin{tcolorbox}[title={Evaluation Prompt Template for HumanEval and HumanEval+}, colback=red!0, left=2pt,right=2pt,top=2pt,bottom=2pt]

{
  Here is a problem for which you need to complete code:

\vspace{0.3cm}
\{instruction\}
\vspace{0.3cm}

Please continue to complete the code with python programming language.

\vspace{0.3cm}
The solution should be in the following format:

\vspace{0.3cm}

  ```python

\vspace{0.3cm}

  \# Your code here

\vspace{0.3cm}

  ```

\vspace{0.3cm}

    Do not generate any tests. You are not allowed to modify the given code and do the completion only.
}
\end{tcolorbox}

  \caption{Prompt template for code evaluation on HumanEval and HumanEval+}
  \label{fig:he_prompt}
  \vspace{-0.2in}
\end{figure}

\label{appendix:G}

\end{document}